\documentclass[conference]{IEEEtran}
\IEEEoverridecommandlockouts
\usepackage{cite}
\usepackage{amsmath,amssymb,amsfonts}

\usepackage{graphicx}
\usepackage{textcomp}
\usepackage{xcolor}
\usepackage{subfig}
\usepackage{algorithm}
\usepackage{algorithmic}

\usepackage{amsmath}
\usepackage{algorithm}

\def\BibTeX{{\rm B\kern-.05em{\sc i\kern-.025em b}\kern-.08em
    T\kern-.1667em\lower.7ex\hbox{E}\kern-.125emX}}
\begin{document}

\title{Dynamic Graph Embedding via LSTM History Tracking}

\author{\IEEEauthorblockN{Shima Khoshraftar\IEEEauthorrefmark{1},
Sedigheh Mahdavi\IEEEauthorrefmark{1},
Aijun An\IEEEauthorrefmark{1}, 
Yonggang Hu\IEEEauthorrefmark{2} and
Junfeng Liu\IEEEauthorrefmark{2}}
\IEEEauthorblockA{\IEEEauthorrefmark{1} Electrical Engineering and Computer Science Department\\
York University,
Toronto, Canada\\ Email: \{khoshraf, smahdavi, aan\}@eecs.yorku.ca}
\IEEEauthorblockA{\IEEEauthorrefmark{2}IBM, Toronto, Canada\\
Email: \{yhu, jfliu\}@ca.ibm.com}
}

\maketitle

\begin{abstract}
Many real world networks are very large and constantly change over time. These dynamic networks exist in various domains such as social networks, traffic networks and biological interactions. To handle large dynamic networks in downstream applications such as link prediction and anomaly detection, it is essential for such networks to be transferred into a low dimensional space. Recently, network embedding, a technique that converts a large graph into a low-dimensional representation, has become increasingly popular due to its strength in preserving the structure of a network. Efficient dynamic network embedding, however, has not yet been fully explored. In this paper, we present a dynamic network embedding method that integrates the history of nodes over time into the current state of nodes. The key contribution of our work is 1) generating dynamic network embedding by combining both dynamic and static node information 2) tracking history of neighbors of nodes using LSTM 3) significantly decreasing the time and memory by training an autoencoder LSTM model using temporal walks rather than adjacency matrices of graphs which are the common practice.  We evaluate our method in multiple applications such as anomaly detection, link prediction and node classification in datasets from various domains.

\end{abstract}


\section{Introduction}
Graphs are powerful tools to represent many of the real world data such as social networks, protein-protein networks, traffic data and scientific collaborations. For instance, in social networks, nodes of the graph are people and connections between them are the edges of the graph. Graphs can be either static or dynamic. In static graphs, the structure of graph is fixed but in dynamic graphs nodes and edges get added and deleted over time.

Many graph mining algorithms deal with static graphs. With the growing number of dynamic networks in the real world, there is a tremendous need for developing efficient algorithms that work properly in dynamic settings. Furthermore, the size of a network increases rapidly over time and makes it challenging to have a proper representation of the entire data. This creates the need for effective algorithms that map networks into a low dimensional space so that they can be utilized in downstream machine learning applications.

A widely used tool to represent graphs are adjacency matrices. The problem with adjacency matrices is that they are memory intensive for very large graphs. Various algorithms were developed to lower the dimensions of these matrices while preserving the necessary information. These algorithms include matrix factorization methods \cite{ahmed2013distributed, belkin2003laplacian,cao2015grarep, tang2011leveraging} and linear or non-linear dimensionality reduction techniques such as Principal Component Analysis (PCA) \cite{wold1987principal}. In general, these methods are computationally expensive for large graphs and fail to perform well in many graph mining tasks.

A recent alternative approach to learning node embeddings is deep learning techniques designed for static graphs. The state-of-the-art static methods are DeepWalk \cite{perozzi2014deepwalk}, LINE\cite{tang2015line}, node2vec \cite{grover2016node2vec} and Variational graph autoencoder \cite{kipf2016variational}. DeepWalk\cite{perozzi2014deepwalk} and Node2vec \cite{grover2016node2vec} take advantage of random walk for each node to preserve the neighborhood structure of the graph. Compared to other embedding methods that use adjacency matrices as input, random walk based methods capture more than immediate neighborhood of nodes. As a result, they can produce more accurate representative vectors for nodes. Furthermore, training the neural networks with random walks are significantly faster than adjacency matrices based methods. 

However, it is still a challenging problem to apply deep node embedding methods for dynamic graphs. Dynamic graphs can change with different rates over time and we need mechanisms to reflect temporal changes in the node embeddings. There have been some recent efforts in this direction that resulted in several algorithms \cite{goyal2018dyngem, nguyen2018continuous, mahdavi2018dynnode2vec, trivedi2018representation, kipf2016variational}. DynGEM \cite{goyal2018dyngem} initializes its model with embeddings from previous time points to stabilize the overall embeddings in consecutive time points.  The difficulty with this method is that it works with adjacency matrices which makes it an memory intensive method. The method in \cite{nguyen2018continuous} generates dynamic embeddings for each node by defining temporal walks for the nodes and utilizes them in training a skipgram model \cite{mikolov2013efficient}, which does not consider the order of the elements in the temporal walks. However, it is essential to avoid losing the order of time dependencies among sequence elements. On the other hand, LSTM proved effective in word and sentence representations in natural language processing. It is specifically useful in preserving long term dependencies between elements in a sequence \cite{wang2016learning, li2015hierarchical,zhang2018sentence}.

In this work, we propose LSTM-Node2vec for computing node embeddings in dynamic graphs. Our key contribution is using an autoencoder LSTM for keeping the history of nodes and training it with a special kind of temporal random walks that capture the evolving patterns in the structure of the graphs. In our method, the embeddings obtained from history are used as initial weights for a node2vec model. Afterwards, node2vec considers the local information from the current graph and produces an embedding that is the combination of both temporal and static information for the nodes of the graph. In addition, for aligning the node embeddings over time, the weights of the model at previous time points is passed along to the model for the next graphs. 

In our experiments, we evaluate the performance of LSTM-Node2vec in
11
 anomaly detection, link prediction and node classification on datasets from various domains. Our approach outperforms existing methods in majority of the cases. Overall, our contributions can be summarized as follows:
\begin{itemize}
    \item We propose LSTM-node2vec, a novel dynamic embedding method that captures temporal changes with LSTM and then the learned parameters are transferred into node2vec to incorporate the local structure of each graph.
    \item We train an autoencoder LSTM model with temporal walks to capture the history of nodes over time which is the first study to consider temporal walks as the input to an LSTM model.
    \item We evaluate our method on three main data mining tasks including anomaly detection, link prediction and node classification.
\end{itemize}
The paper is organized as follows. We first explain the previous related works. Then we present the technical details of our method in the next section and show the experiments results next. Finally, in the conclusion section we summarize the findings in this work.

\section{Related Work}
With the emergence of massive networks, the automation of feature engineering for graphs has become an essential task in the research community. The early studies in this field are mainly on dimensionality reduction methods that work on adjacency matrices and Laplacian of graphs \cite{tang2011leveraging, belkin2003laplacian}. These methods have certain drawbacks in terms of time and memory efficiency. Recently, representational learning has breakthroughs in natural language processing \cite{mikolov2013efficient}. These advancements have been transferred into the network mining area in node2vec \cite{grover2016node2vec} and DeepWalk \cite{perozzi2014deepwalk} using the analogy between a document and network, in which nodes in a network are modeled similarly to words in a document based on the idea that words/nodes in the same neighborhood tend to be similar. These methods differ in the sampling strategy chosen to explore the neighbors of a node. DeepWalk\cite{perozzi2014deepwalk}, Node2vec \cite{grover2016node2vec} and LINE \cite{tang2015line} are state-of-the-art random walk based models that use uniform, BFS/DFS like random walks and first/second order distance for this purpose, respectively. Examples of other static node embedding methods are proposed in \cite{donnat2018learning, ribeiro2017struc2vec}.


All the previous methods mentioned above are designed for static graphs. However, as most real world graphs are dynamic, there have been a new set of work on dynamic network embedding. DynGEM \cite{goyal2018dyngem} is a neural net dynamic embedding method that models the graph stream as a series of graph snapshots and initializes the model for each graph using the previous graph weights to stabilize the embedding vectors. CTDN \cite{nguyen2018continuous} learns a continuous-time network representation by capturing temporal dependencies of the networks through random walks. HTNE \cite{zuo2018embedding} is based on Hawkes process and uses a neighborhood formation sequence of a dynamic network to generate dynamic network embeddings. dyngraph2vecAE and dyngraph2vecAERNN are variants of dyngraph2vec method \cite{goyal2018dynamicgem}. These methods are deep learning methods that take a series of graphs and predict the graph at next time point. In order to capture connection between nodes over time, dyngraph2vec uses multiple fully connected layers and dyngraph2vecAERNN is based on LSTM with node representations as inputs. The problem with dyngraph2vecAERNN is that although it decrease memory requirements compare to using adjacency matrices but it has the overhead of computing initial representations first. Our work utilizes random walks to train an LSTM autoencoder to capture dynamics of the networks, which significantly reduces the memory requirements compared to previous methods based on adjacency matrices \cite{goyal2018dyngem, goyal2018dynamicgem}. Other dynamic network embedding methods are presented in \cite{mahdavi2018dynnode2vec, trivedi2018representation, zhou2018dynamic}. 

Network embedding has many applications in link prediction, node classification and anomaly detection. DynGEM \cite{goyal2018dyngem} informally applied node embeddings to change detection by defining the change as the difference in embeddings between consecutive time points. NetWalk \cite{yu2018netwalk} is the first serious attempt to propose a dynamic network embedding method based on deep neural networks for anomaly detection. This method presents a clique embedding technique, uses reservoir sampling to update network representations as new objects arrive, and detect anomalies by a dynamic clustering algorithm. Generally, there is a large body of work on anomaly detection and there is an opportunity to apply deep neural network based methods in this area. Our model, LSTM-Node2vec, shows a significant success in detecting star-shaped anomalies in data. LSTM-Node2vec is specially applicable in anomaly detection because it can identify changes in the history of a node and incorporate it into the node embedding vectors.


\section{Dynamic Network Embedding Method}
\subsection{Problem Statement}
A graph stream is represented by $G_1, G_2,..., G_t$ where $G_i= (V_i, E_i)$ is the graph at time $i$ and $V_i$ and $E_i$ are the nodes and edges of the graph. Given a graph stream, our goal is to compute the dynamic representation vectors for the nodes of graphs at each time point. The embedding of a node $v$ into a $d$-dimensional vector can be formulated as a mapping function $f:v\rightarrow{R^d}$. 

In order to generate an embedding vector for each node in a graph, we propose LSTM-Node2vec that combines both dynamic and static states of a node to generate a more accurate representation of the node.

\subsection{Overview of LSTM-Node2vec}
LSTM-Node2vec generates the node embeddings for graph $G_i$ in the graph stream using three main steps: (1) generating temporal random walks over nodes from a sequence of graphs before $G_i$, (2) training an autoencoder LSTM with the temporal neighbor walks to learn node embedding for $G_i$, and (3) passing the embeddings as initial weights to a node2vec model. Algorithm \ref{alg1} presents the steps of LSTM-Node2vec. Below, we will describe the three steps in detail. 

\begin{algorithm}
\caption{:\textbf{LSTM-Node2vec}}
\label{alg1}
\begin{algorithmic} [1]
\STATE \textbf{Input}: Graphs $G_{t-(L-1)}, G_{t-(L-2)}, ..., G_t$, where $L$ is the temporal window size and $t$ is the current time point. 
\STATE \textbf{Output}: $Z_t$: embedding vectors for all the nodes in $G_t$
\FOR {$i =1$ to $|V_{G_{t}}|$}
\STATE Generate $k$ temporal random walks of maximum length $L$ for node $v_i$ of $G_t$, and add them to set $R$
\ENDFOR
\STATE Initialize the input layer weights of an LSTM encoder with $Z_{t-1}$

\STATE Train the LSTM autoencoder with node sequences in set $R$
\STATE Initialize node2vec with the input layer weights $W_i$ of the trained LSTM encoder
\STATE Train a skipgram model with random walks on $G_t$ 
\STATE Return the weights in the input layer of the skipgram model as $Z_t$
\end{algorithmic}
\end{algorithm}

\subsection{Temporal neighbor walk generation.} 
Inspired by \cite{nguyen2018continuous} and \cite{zuo2018embedding}, we define a temporal random walk as follows.
Given a series of graphs $G_1, G_2, ..., G_t$ a temporal neighbor walk for each node $v$ in graph $G_i$ is defined as a sequence of neighbors of a node $v$ at $L$ previous time points represented by ${u_{i-L}, u_{i-L+1},...,u_i}$ where $u_x$ is the neighbor of node $v$ at time $x$. 

We sample a neighbor for node $v$ at each of $L$ previous time points based on the alias sampling method used in node2vec \cite{grover2016node2vec} for random walk generation. We utilized the same approach (described in Section E below) to sample a neighbor node for node $v$ in each time point. The temporal walk is the concatenation of neighbors of node $v$ in the increasing order of time. By defining the temporal random walk this way, we track the changes of neighbors of a node at each time point and as a result the random walk reflects the changes in the structure of the graph over time. For instance, in the coauthorship temporal network, the temporal walk shows the changes in the collaboration of a particular author over time. Figure \ref{fig3} illustrate the temporal walk generation in a temporal network for window size $L=3$. 

The window size that we consider for generating temporal walk is $L$. The length of the temporal walk is allowed to have a value between $[2, L]$. 2 is the minimum required length to define a valid sequence and $L$ is the maximum window size. As mentioned in \cite{nguyen2018continuous}, it is possible that a node does not occur in several time points in the $L$ previous time points. As a result, depending of how frequent a node is in the window, the length of its temporal walk ranges between $2$ and $L$. In addition, we generate $k$ temporal walks per node in each graph. Therefore, the total number of generated temporal walks is a multiple of the total number of nodes in the graph.






\begin{figure}
    \centering
    \includegraphics[scale=0.5]{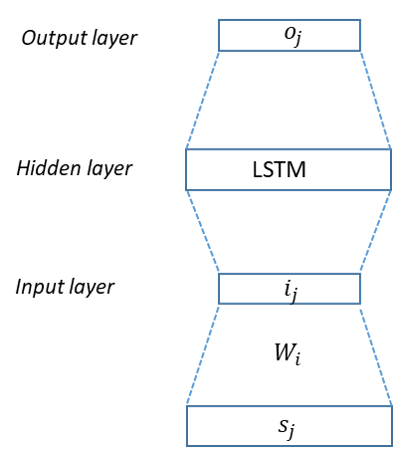}
    \caption{LSTM training for word representation. $s_j$ is the $j$th word in a sentence which is represented by a one-hot encoding vector. $W_i$ is the weights of input layer in LSTM.  
    }
    \label{fig2}
\end{figure}

\begin{figure*}
    \centering
    \includegraphics[scale=0.5]{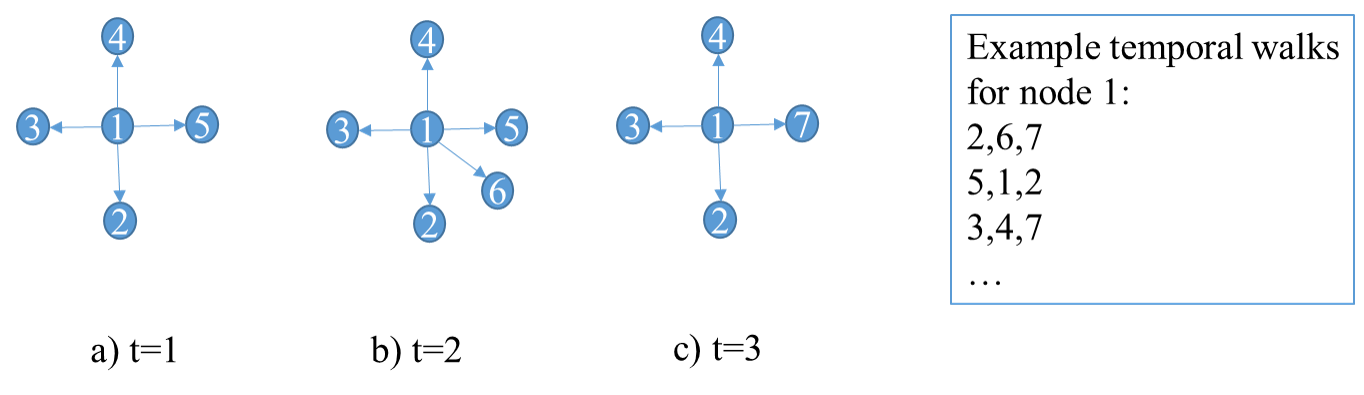}
    \caption{The process of temporal walk generation. a, b, c are three time points of a temporal network. Given $L=3$ the temporal walks for node 1 is generated by sampling from its neighborhood nodes. }
    \label{fig3}
\end{figure*}

\subsection{LSTM Autoencoder}
An LSTM is a recurrent neural network (RNN) used for capturing long term order dependencies between elements in sequences. LSTMs are widely used in many areas of machine learning, specially natural language processing (NLP), for problems such as text translation and question answering \cite{sutskever2014sequence, du2017identifying, chen2018learningq, wang2018qg, zhao2018paragraph, koehn2007moses, bahdanau2014neural}. 
A common usage of LSTM in NLP is to generate word representations \cite{wang2016learning,li2015hierarchical,zhang2018sentence}. Given a set of sentences, an LSTM autoencoder is trained to take a sentence as input and outputs the same sentence. Given a sentence $S=\{s_1, s_2, ..., s_n\}$ of length $n$ where $s_i$ is the $i$th word in the sentence, an LSTM autoencoder takes the words one by one using the one-hot word representation and is trained to produce the same word in the output layer. After the training, the weights in the input layers are the word representation for $s_i$ is obtained from learned weights $W_i$ in LSTM. Figure \ref{fig2} depicts the LSTM training for word representations.

Inspired by the LSTM effectiveness in NLP, we use LSTM to learn node representations in graphs. Here, a temporal walk over nodes at different time points is considered as a sentence. The order in the temporal walk reflects the way that the neighborhood of a node evolves over time. By training an LSTM over a set of temporal random walks, LSTM can capture the temporal dependencies between the neighbors of nodes in the graph evolution. Let $R$ denote a set of all temporal walks obtained for nodes of graph $G_t$, $R=\{r_1, r_2, ..., r_m\}$. $r_i={u_1, u_2,...,u_L}$ is a temporal walk generated for each node in the graph. $m = N \times k$ is the total number of walks when there is $N$ nodes and $k$ temporal walks are generated for each node. We used each $r_i$ in $R$ for training LSTM autoencoder such that $r_i$ is the input and output of the model. The formula of LSTM in the encoder part of the model is as follows: 
\begin{align*}
    i_t&=\sigma(W_i[h_{t-1},u_t]+b_i)\\
    f_t&=\sigma(W_f[h_{t-1},u_t]+b_f)\\
    o_t&=\sigma(W_o[h_{t-1},u_t]+b_o)\\
    \Tilde{C_t}&=tanh(W_c[h_{t-1},u_t]+b_c)\\
    C_t &= f_t \times C_{t-1}+i_t \times \Tilde{C_t}\\
    h_t &= o_t \times tanh(C_t)
\end{align*}

In these equations, $f_t,i_t,o_t$ are forget gate, input gate and output gate. $h_{t-1}$ is the output of LSTM at previous time point. $C_t$ is the cell state vector which is updated in two parts. First forget gate $f_t$ decides which part of cell state in previous time $C_{t-1}$ will be discarded and then the new values will be stored in cell state. The output of LSTM in the current time point $h_t$ will be a filtered version of cell state. $b_x, W_x$ are biases and weights for respective gates. $u_t$ is each node in the temporal walk and $W_i$ are the node representations.

We run LSTM-Node2vec for every graph in the stream to get the node embeddings of the graph. It is crucial that the generated embeddings for consecutive graphs be in the same vector space. In order to do that, the LSTM autoenceder for the first graph $G_0$ is randomly initialized. Then, we initialize the LSTM autoencoder of $G_{i}$ with the weights of the previous graph $G_{i-1}$ model. 


\subsection{Node2vec}
Node2vec \cite{grover2016node2vec} is the state-of-the-art node representation method for static graphs. The node2vec algorithm is built on the concepts that were first introduced in DeepWalk algorithm. Both of these methods are categorized as random walk based low dimensional node representation algorithms. Node2vec algorithm consists of two parts: random walk generation and skipgram \cite{mikolov2013efficient}. The main idea of node2vec is to generate second order random walks for all nodes in the graph such that it effectively explore neighborhood of the node by interpolating between DFS and BFS search strategies. In order to create that effect, node2vec defines a biased random walk for each node using two parameters $p$ and $q$ that control the random walk exploration procedure. Specifically, $p$ is a return parameter which can change the probability that the walk returns to an already visited node. $q$ is the inward and outward exploration parameter. Large values of $q$ guide the walk toward exploration of nodes that are close to the target which is similar to how BFS works and small values of $q$ encourage outward exploration similar to DFS. 
Formally the unnormalized transition probability between node $v$ and node $u$ is defined as follows: 
\begin{equation}
\alpha_{pq}(v,u)=
\begin{cases}
 1/p, &\text{if $d_{vu} = 0$}\\
 1, &\text{if $d_{vu} = 1$}\\
 1/q, &\text{ if $d_{vu} = 2$}
\end{cases}
\end{equation}
where $d_{vu}$ is the shortest path distance between nodes $v$ and $u$. 


In the next step of node2vec, the random walks generated in previous step are used to train a skipgram architecture. Skipgram model learns the continuous representations of each node. The objective function of node2vec is a maximum likelihood optimization problem that maximizes the probability of preserving neighborhood of a node in a d-dimensional space. Here is the objective function formula:
\begin{equation}
    max \sum_{u\in V}{log Pr(N_s(u)|f(u))}
\end{equation}
where $u$ is the target node in the graph $G=(V,E)$. $N_s(u)$ and $f(u)$ are the sample neighborhood of node $u$ generated using random walks and the embedding vector of $u$, respectively. In the context of Natural Language Processing, neighborhood of a word in a sentence is defined as a window of words close to the target word in the sentence. This concept is transferred to graphs using random walks that depending on the definition sample different neighbors of a node. 

We initialize the embedding layer of the node2vec skipgram model with the learned weights in the input layer of LSTM autoencoder for each graph at time $t$. This way, the final node representations in node2vec for graph $G_t$, which are the output of LSTM-Node2vec, are the combination of both static and dynamic states of the graph.





\begin{figure*}
    \centering
    \includegraphics[scale=0.5]{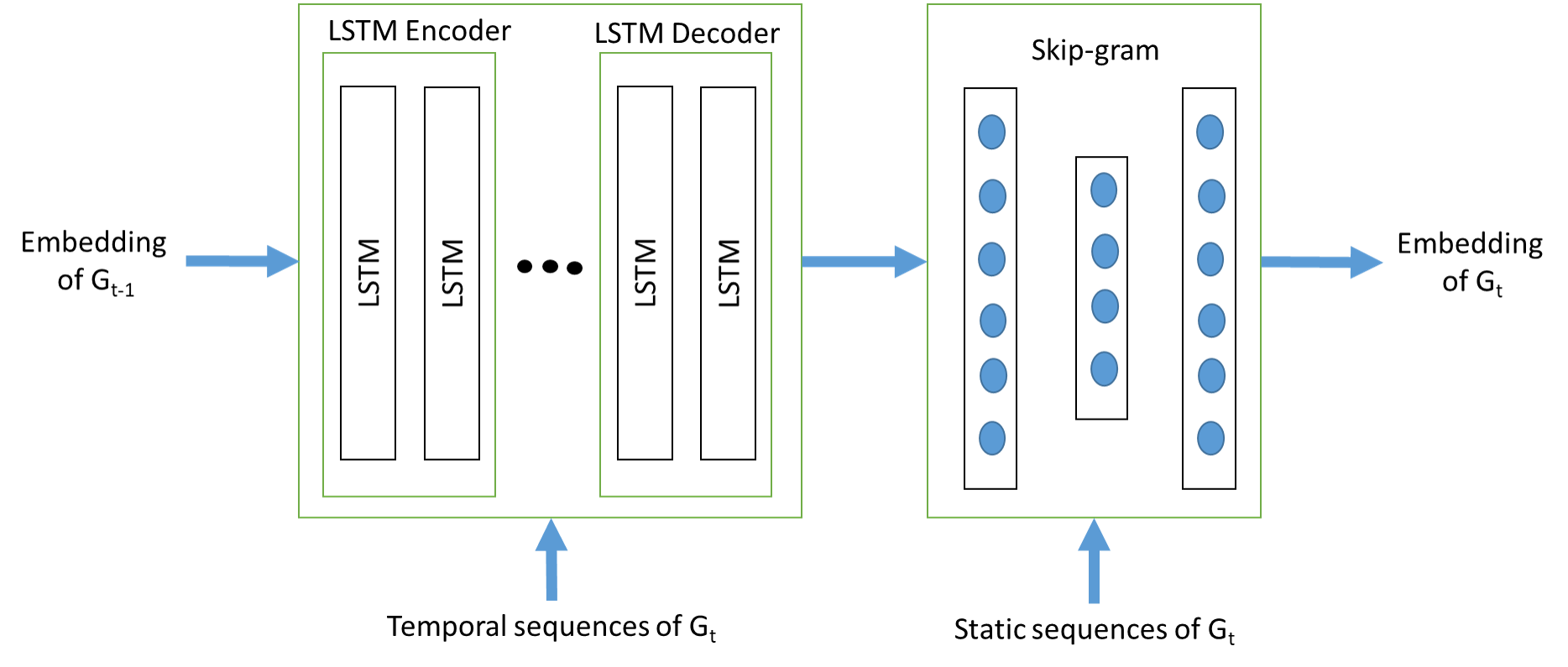}
    \caption{Framework for LSTM-Node2vec for creating embedding of $G_t$ using both temporal and static sequences}
    \label{fig1}
\end{figure*}

\section{Experiments}
We applied the generated node representation vectors in different data mining tasks in different datasets. The results of the experiments are given below. 
\subsection{Baselines}
The evaluation of LSTM-Node2vec was performed against following the state-of-art methods for static and dynamic node representations.

\begin{itemize}
    \item \textbf{DeepWalk \cite{perozzi2014deepwalk}:} DeepWalk is a static network embedding method based on random walks that utilize uniform random walks.
    \item \textbf{Node2vec \cite{grover2016node2vec}:} This method is a static network representation algorithm consisting of BFS/DFS like random walks and skipgram.
    \item \textbf{dyngraph2vecAE \cite{goyal2018dynamicgem}:} It is a dynamic network embedding method based on dyngraph2vec. It utilizes deep learning models with multiple fully connected layers to model the interconnection of nodes.
    \item \textbf{dyngraph2vecAERNN \cite{goyal2018dynamicgem}:} This is another variant of the dyngraph2vec method which is a dynamic representation learning method. It feeds previously learned representations to LSTMs to generates embedding vectors.
\end{itemize}
\subsection{Experiment settings}
In all the experiments the embedding vector size is 128. The parameter $L$ in LSTM-Node2vec is selected depending on the size of each dataset. For Radoslaw, Ubuntu, Contact, St-Ov, AS, Dblp and Acm, the length of $L$ equals 10, 10, 10, 20, 20, 5 and 5, respectively. The LSTM-Node2vec model is trained with the Adam optimizer. We run Node2vec with $(p,q)=(0.25,1)$ and DeepWalk with $(p,q)=(1,1)$.

\subsection{Anomaly Detection}
Anomaly detection is an important data mining and graph mining application. Anomalies are any deviation from normal behavior. Identifying these irregular patterns in the data is the task of anomaly detection methods \cite{ranshous2015anomaly}. Anomaly detection is widely applied in static and dynamic network mining tasks such as network intrusion detection, bank frauds and social networks \cite{ding2012intrusion, xu2010intrusion,ghoting2004loaded,zhang2005anomalous}. Comprehensive description of anomaly detection methods in graphs can be found in two surveys \cite{ranshous2015anomaly, akoglu2015graph}.  We evaluated the performance of LSTM-Node2vec in anomaly detection task and compared the results with those of other static and dynamic embedding methods.



\subsubsection{Datasets}
We run our experiments on the following three datasets:
\begin{itemize}
    \item Radoslaw\cite{nr}: This is a dataset of email communications between employees of a company with 167 nodes, 89K edges over 39 time points.
    
    \item Ubuntu\cite{nr}: Ubuntu is a network of interactions on the Ask Ubuntu website. Interaction between users include answering/commenting on other users questions/comments. This temporal dataset consists of 137K nodes, 280K edges and has 79 time steps.  
    
    \item Contact\cite{nr}: This dataset is a network of human contacts with 274 nodes and 28.2K edges. We divided the dataset into 39 graphs.
\end{itemize}
\begin{table}[]
\centering
\caption {AUC results of anomaly detection} \label{tab1}
\begin{tabular}{l|lll}
   Algorithm                    & \multicolumn{1}{l|}{Radoslaw} & \multicolumn{1}{l|}{Ubuntu}   & Contact \\ \hline
DeepWalk               & \multicolumn{1}{l|}{0.72095}  & \multicolumn{1}{l|}{0.53984}  & 0.67082  \\
node2vec               & \multicolumn{1}{l|}{0.715568} & \multicolumn{1}{l|}{0.674564}  &      0.64213    \\
dyngraph2vecAE         & \multicolumn{1}{l|}{0.431883} & \multicolumn{1}{l|}{0.5787}         & 0.732996 \\
dyngraph2vecAERNN      & \multicolumn{1}{l|}{0.477085} & \multicolumn{1}{l|}{0.5965}         & \textbf{0.76226}  \\ \hline
\textbf{LSTM-Node2vec} & \multicolumn{1}{l|}{\textbf{0.895336}} & \multicolumn{1}{l|}{\textbf{0.703182}} & 0.682581 \\ 
\end{tabular}
\end{table}

\subsubsection{Injecting anomalies}
Because of the challenges in finding datasets that have ground truth labels in anomaly detection, we directly injected anomalies into normal datasets \cite{akoglu2015graph, yu2018netwalk}. In network intrusion detection, one of the important type of anomalies are bursts in activity. For instance, bursts can happen when a particular node in the network starts attacking other nodes. This creates a star shaped dense subgraph in the graph structure. 

In this work, we inject star shaped anomalies into three datasets. For creating star anomalies, we add a large number of new edges between a target node and $n$ other existing nodes in the dataset that did not have an edge with the target node previously. This shows a sudden burst of activity from the target node toward other nodes. We create multiple star shaped anomalies and then inject them into dynamic graphs in each dataset. In order to do that, we inject one anomaly in $k$ consecutive graphs. For instance, for $k=3$ in graph streams $G_1, G_2, ..., G_{10}$ we inject anomalies in $G_3, G_4, G_5$. Then we jump $m$ graphs and start injecting the second anomaly and continue similarly on the entire graphs in the dataset.

\begin{figure}
    \centering
    \includegraphics[scale=0.5]{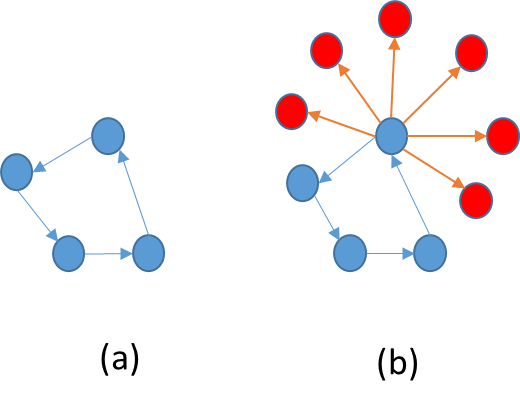}
    \caption{A temporal network at two time steps. (a) the graph at time $t-1$. (b) the graph at time $t$ with an star shaped anomaly, one previously normal node starts attacking multiple nodes shown in red.}
    \label{fig1}
\end{figure}

\subsubsection{Classification}
We use the anomalous datasets generated in previous step for anomaly detection tasks. Here, we formulate the anomaly detection as a classification task. Consider a graph stream of $G_1, G_2, ..., G_t$. First, the node representation vector of all the nodes in each graph in the stream is computed. In our datasets, edges have anomalous or normal class labels because each edge represents an attack from a source node to a destination node. As a result, we create edge embedding for all the edges in the stream. There are multiple known ways to create edge embeddings from node embedding vectors including Hadamard, $l_1, l_2$ and average \cite{grover2016node2vec}. In our experiments, we used $l_1$ operator meaning that for an edge $(u,v)$, $f(u,v)=f(u)-f(v)$, where $f(x)$ is the embedding of $x$.

After creating edge embeddings for each graph, we perform classification on graph $G_i$ using edge embeddings of graphs in $G_0, G_1,..., G_{i-1}$ for $i$ from $L$ to $t$. The reported value for each dataset is the average AUC score of classification task over the entire time points using the Random Forest classifier. The results in Table \ref{tab1} show that LSTM-Node2vec outperforms other embedding methods in Radoslaw and Ubuntu datasets and the average AUC score in Contact is less than two other methods. This can be due to the fact that the changes in the structure of graphs in Radoslaw and Ubuntu are rather smoother than Contacts and the history is still relevant in the current time. Therefore, we can say that LSTM-Node2vec works well when the changes of graphs are not significant over time.

\begin{table}[]
\centering
\caption{Anomaly detection AUC scores for three graphs to analyze the effect of previously seen anomalies }
\label{tab2}
\begin{tabular}{l|lll}
 & $G_1$ & $G_2$ & $G_3$ \\ \hline
Radoslaw & 0.88 & 0.90 & 0.91 \\
Ubuntu & 0.64 & 0.89 & 0.97 \\
Contact & 0.76 & 0.95 & 0.99
\end{tabular}
\end{table}

\subsubsection{Detecting previously seen anomalies}
It is expected that the anomaly detection accuracy increases if the anomalies have previously occurred in the dataset. We tested this on three datasets. As explained in previous sections, one anomaly is injected into $k=3$ consecutive time points and the AUC score of anomaly detection in each of anomalous graphs is computed. The results are shown in Table \ref{tab2}. We first introduced the anomaly in graph $G_1$. As a result, it has the lowest anomaly detection score. For $G_2$ and $G_3$, however, as the anomaly has existed in the dataset, the accuracy starts to increase.

\begin{table}[]
\centering
\caption {Macro-f1 and Micro-f1 scores for node classification } \label{tab3}
\begin{tabular}{l|l|ll}
Metric   & Algorithm          & \multicolumn{1}{l|}{Dblp}   & Acm    \\ \hline
Macro-f1 & DeepWalk               & \multicolumn{1}{l|}{0.34}   & 0.3458 \\
         & node2vec               & \multicolumn{1}{l|}{0.3333}       &  0.3696      \\
         & dyngraph2vecAE         & \multicolumn{1}{l|}{0.3299}       &        0.334\\
         & dyngraph2vecAERNN      & \multicolumn{1}{l|}{0.3682}       &        0.3969\\ \cline{2-4} 
         & \textbf{LSTM-Node2vec} & \multicolumn{1}{l|}{\textbf{0.4658}} & \textbf{0.4605} \\ \hline
Micro-f1 & DeepWalk               & \multicolumn{1}{l|}{0.4632} & 0.4908 \\
         & node2vec               & \multicolumn{1}{l|}{0.4624}& \textbf{0.5291}        \\
         & dyngraph2vecAE         & \multicolumn{1}{l|}{0.4523}       &        0.4656\\
         & dyngraph2vecAERNN      & \multicolumn{1}{l|}{0.4645}       &        0.4962\\ \cline{2-4} 
         & \textbf{LSTM-Node2vec} & \multicolumn{1}{l|}{\textbf{0.501}} & 0.4995  \\ 
\end{tabular}
\end{table}

\subsection{Node Classification}
Node classification is to classify a node in a graph into a predefined category.
Similar to any classification task, part of the dataset is considered as a training set and the class labels for the test set are predicted.  An approach for node classification is presented in \cite{goyal2018dyngem} in the dynamic graph setting. Based on this approach, we classified each graph at time $t$ using the previous graph at $t-1$ by applying the logistic regression method.  We used two measures Macro-f1 and Micro-f1 for evaluating our method. The results are reported in Table \ref{tab3}. The datasets used are as follows:
\begin{itemize}
    \item Dblp\cite{tang2008extraction,tang2008arnetminer}: Dblp is the network of coauthorship between researchers from 2000 to 2017 with 90k nodes and 749k edges. Nodes in this dataset are the researchers that belong to either database/data mining class (VLDB, SIGMOD, PODS, ICDE, EDBT, SIGKDD, ICDM, DASFAA, SSDBM, CIKM, PAKDD, PKDD, SDM and DEXA) or computer vision/pattern recognition class (CVPR, ICCV, ICIP, ICPR, ECCV, ICME and ACM-MM).
    \item Acm\cite{tang2008extraction,tang2008arnetminer}: The Acm dataset is similar to Dblp and spans from 2000 to 2015.
\end{itemize}

As it is evident in Table \ref{tab3}, LSTM-Node2vec outperforms other methods in the Dblp dataset in terms of both Macro-f1 and Micro-f1. In Acm, our performance gain is above other methods in terms of Marcro-f1 and it is comparable with other results in terms of Micro-f1. LSTM-Node2vec is remarkable in the node classification task. This is reasonable as a class label in Acm and Dblp datasets is an area of research of authors which does not drastically change over time. Thus, including history in the embedding computation helps in giving a better picture of the author in the current time.

\begin{figure*}[!ht]
   \centering
   \subfloat[][Radoslaw]{\includegraphics[width=.3\textwidth]{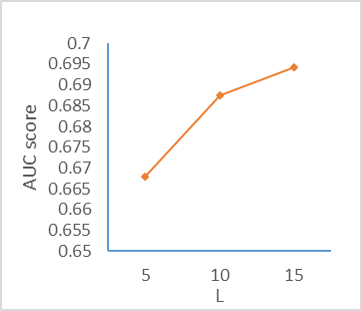}}\quad
   \subfloat[][St-Ov]{\includegraphics[width=.3\textwidth]{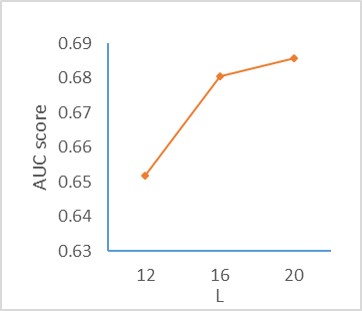}}\\
   \caption{Analysis of effect of parameter $L$ on link prediction for Radoslaw and St-Ov datasets}
   \label{fig:sub1}
\end{figure*}

\begin{figure*}[!ht]
   \centering
   \subfloat[][Acm]{\includegraphics[width=.3\textwidth]{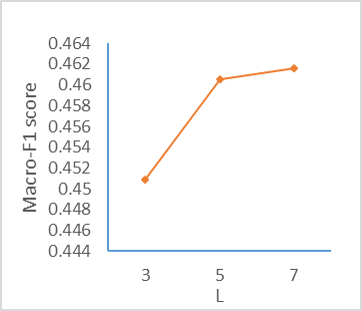}}\quad
   \subfloat[][Dblp]{\includegraphics[width=.3\textwidth]{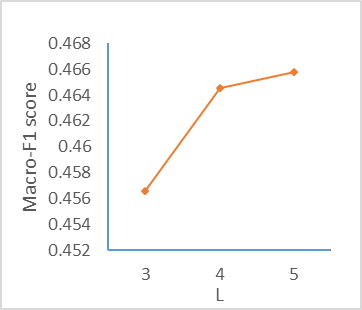}}\\
   \caption{Analysis of effect of parameter $L$ on node classification for Acm and Dblp datasets}
   \label{fig:sub2}
\end{figure*}

\subsection{Link Prediction}
One of the main graph mining tasks is link prediction. Link prediction can be formulated as a classification task such that each edge either has a positive or negative class label. Then the edges in the test set are classified using a model trained with the training set. For this purpose, we predict edges in graph $G_t$ using previous graphs from $0$ to $t-1$ in the graph stream, based on a method proposed in \cite{goyal2018dyngem}. Our classifier is logistic regression. Table \ref{tab4} summarizes the results in terms of the average AUC score for three datasets. This evaluation is performed on the following three dataset.
\begin{itemize}
    \item St-Ov\cite{leskovec2015snap}: This is the user interaction network in the Math Overflow website. St-Ov contains 14k nodes and 195k edges with 58 time steps.
    \item Radoslaw\cite{nr}: Explained in the anomaly detection section.
    \item AS\cite{leskovec2015snap}: AS is the communication network of who-talk-to-whom gathered from logs of Border Gateway Protocol. The numbers of nodes, edges and time steps are 6k, 13k and 100 respectively.
\end{itemize}

Table \ref{tab4} shows that LSTM-Node2vec outperforms other methods in terms of AUC scores on all the datasets. Node2vec also has a good performance compared to other methods. LSTM-Node2vec improves that even more by including the temporal information of nodes.

\begin{table}[]
\centering
\caption {AUC scores for link prediction} \label{tab4}
\begin{tabular}{l|lll}
Algorithm                        & \multicolumn{1}{l|}{St-Ov} & \multicolumn{1}{l|}{Radoslaw} & AS     \\ \hline
DeepWalk               & \multicolumn{1}{l|}{0.5976}         & \multicolumn{1}{l|}{0.6516}   & 0.8858 \\
node2vec               & \multicolumn{1}{l|}{0.6038}       & \multicolumn{1}{l|}{0.6738}   &    0.8577
    \\
dyngraph2vecAE         & \multicolumn{1}{l|}{0.5098}               & \multicolumn{1}{l|}{0.5233}         &     0.7193   \\
dyngraph2vecAERNN      & \multicolumn{1}{l|}{0.5372}               & \multicolumn{1}{l|}{0.5612}         &        0.7085\\ \hline
\textbf{LSTM-Node2vec} & \multicolumn{1}{l|}{\textbf{0.6857}}         & \multicolumn{1}{l|}{\textbf{0.6875}}   & \textbf{0.8878} \\ 
\end{tabular}
\end{table}

\subsection{Effect of length of history parameter $L$}
We analyze the influence of length of $L$ in the link prediction and node classification tasks. We computed the scores using different lengths of $L$ depending on the datasets. For instance, as Dblp consists of 18 graphs, we compute the results for $L=3,4,5$. The same is true for the Acm dataset. We use longer $L$ values for Radoslaw and specially St-Ov as they are bigger datasets. The results of this analysis are given in Figures \ref{fig:sub1} and \ref{fig:sub2}. They show that increasing the length of history has a positive effects in the results. However, based on our observations, there is a limit to this increase which is due to the fact that by including very far history in the current state of a graph, we may consider information that are not relevant to present time.

\subsection{Effects of changes in model structure}
Evaluating LSTM-Node2vec for all three tasks was performed using one LSTM layer in the encoder and one LSTM layer in the decoder. In order to analyze the effect of adding complexity to the model, we computed the results with an additional LSTM layer in the encoder and reported the difference in performance in Table \ref{tab5}. The results show that making the model more complex does not significantly improve the performance of the model and in some cases even negatively effects the results. In general, this means that the LSTM-Node2vec model with one layer for both the encoder and decoder can return results that are comparable to more complex models.

\begin{table}[]
\centering
\caption{Difference in performance by adding one more layer to the encoder}
\label{tab5}
\begin{tabular}{l|lll}
 & AUC & Mac-f1 & Mic-f1 \\ \hline
Radoslaw & -0.0085 & - & - \\
St-Ov & -0.0014 & - & - \\
AS & -0.0096 & - & - \\
Dblp & - & -0.0045 & 0.0086 \\
Acm & - & 0.0036 & 0.0059
\end{tabular}
\end{table}

\subsection{Time Analysis}
\begin{figure}
    \centering
    \includegraphics[scale=0.65]{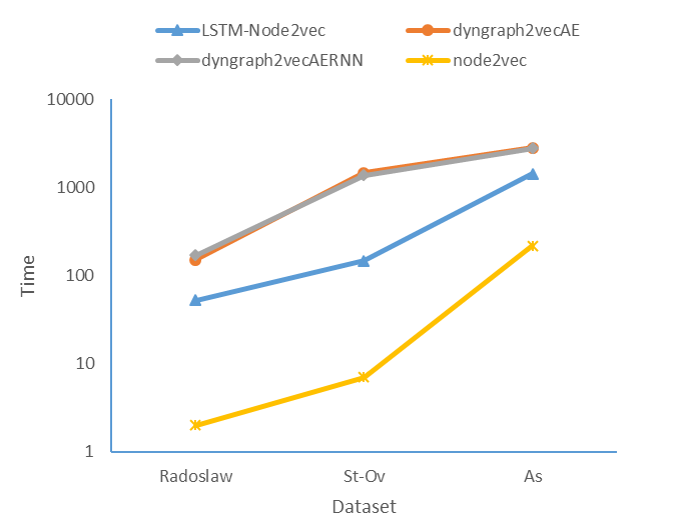}
    \caption{Time complexity Analysis for 4 time steps in 3 datasets}
    \label{fig8}
\end{figure}

We compare the running time of LSTM-Node2vec to the static embedding model Node2vec and two dynamic models dyngraph2vecAE and dyngraph2vecAERNN on three datasets, AS, Radoslaw and St-Ov. We ran the experiments on a Ubuntu server with 512 GB RAM, 7 GPUs with 2 x Intel Xeon E5-2687W v4 3.0 GHz each. The results in Figure \ref{fig8} show that LSTM-Node2vec is slower than Node2vec. This is because Node2vec only focuses on one graph for the computation. However, LSTM-Node2vec uses historical information from previous graphs in the current graph embedding. Compared to the two dynamic embedding models, the running time of LSTM-Node2vec is better as LSTM-Node2vec uses random walks instead of adjacency matrices of the graphs. We are working on implementing a distributed version of LSTM-Node2vec to lower the running time further.

\section{Conclusion}
We have presented LSTM-Node2vec, a dynamic network embedding method that combines an LSTM autoencoder and the node2vec model to learn node representations in dynamic graphs. We defined temporal neighbor walks that capture evolving patterns in the history of nodes of the graphs. In order to generate dynamic network embeddings, we first trained an LSTM autoencder with temporal neighbor walks to extract temporal information of nodes over time. The weights of the LSTM model is then used to initialize a node2vec model to learn node representations based on the structure of the current graph. This way we incorporated temporal information into static local states of the nodes, which leads to producing better node representations. We evaluated the effectiveness of our model in three applications including anomaly detection, link prediction and node classification tasks. Our method outperformed other state-of-the-art static and dynamic embedding methods in most of the cases. For future work, we are interested in adding attention mechanism to the LSTM-Node2vec model. This approach can be specially useful in the anomaly detection application that we want the model to focus more on the anomalous parts of the dataset. Furthermore, in the current version of the method, the length of $L$ is fixed. One interesting future work is to automate the process of choosing the length of $L$. This way the model can adjust $L$ based on how much of the history it wants to include in the computation. Therefore, if the history of a graph is not relevant to the current state of the graph, the effects of it in the current embedding will be limited. Similarly, if a graph does not change significantly over time, the history will be given better consideration. 

\section*{Acknowledgment}
This work is funded by the Natural Sciences and Engineering Research Council of Canada (NSERC), IBM Canada and the Big Data Research Analytics and Information Network (BRAIN) Alliance established by Ontario Research Fund - Research Excellence Program (ORF-RE).






\bibliographystyle{ieeetr}
\bibliography{ijcai19}


\end{document}